\documentclass{ifacconf}
\usepackage{url}
\usepackage{subfig}
\usepackage{booktabs}
\usepackage{graphicx, caption, subcaption}   %
\usepackage{natbib}        %
\usepackage{amsfonts,amsmath,amssymb,mathrsfs} 
\newtheorem{theorem}{Theorem}
\newtheorem{remark}[theorem]{Remark}
\usepackage{tikz}
\usepackage{multicol}
\usepackage{multirow}
\usetikzlibrary{positioning}
\begin{document}
\begin{frontmatter}

\title{Inverse Optimality for Fair Digital Twins: A Preference-based approach}

\author[First]{Daniele Masti} 
\author[First]{Francesco Basciani} 
\author[First]{Arianna Fedeli} 
\author[third]{Giorgio Gnecco} 
\author[Fourth]{Francesco Smarra} 
\address[First]{Gran Sasso Science Institute, 
   Viale F. Crispi 7, L'Aquila, 67100 Italy (e-mail: 
   \{daniele.masti,francesco.basciani,arianna.fedeli\}@gssi.it).}
\address[third]{Scuola IMT Alti Studi Lucca, 
   Piazza San Francesco 19, Lucca, 55100 Italy (e-mail: giorgio.gnecco@imtlucca.it).}
\address[Fourth]{Fondazione Ugo Bordoni, 
   Viale del Policlinico 147,Rome, 00161 Italy (e-mail: fsmarra@fub.it).}

\begin{abstract}
Digital Twins (DTs) are increasingly used as autonomous decision-makers in complex socio-technical systems. However, their mathematically optimal decisions often diverge from human expectations, revealing a persistent mismatch between algorithmic and bounded human rationality. This work addresses this challenge by proposing a framework that introduces fairness as a learnable objective within optimization-based Digital Twins. In this respect, a preference-driven learning workflow that infers latent fairness objectives directly from human pairwise preferences over feasible decisions is introduced. A dedicated Siamese neural network is developed to generate convex quadratic cost functions conditioned on contextual information. The resulting surrogate objectives drive the optimization procedure toward solutions that better reflect human-perceived fairness while maintaining computational efficiency. The effectiveness of the approach is demonstrated on a COVID-19 hospital resource allocation scenario. Overall, this work offers a practical solution to integrate human-centered fairness into the design of autonomous decision-making systems.
\end{abstract}

\begin{keyword}
Human-centered production and logistics; Human-technology integration in manufacturing; Data-driven and AI-based modelling of production and logistics. %
\end{keyword}

\end{frontmatter}
\section{Introduction}

Digital Twins (DTs) are increasingly deployed as autonomous decision-makers in complex socio-technical systems, ranging from energy management~\citep{djebali2024survey} and mobility planning~\citep{somma2025model} to healthcare logistics~\citep{katsoulakis2024digital}. In these settings, the digital counterpart continuously observes, predicts and optimizes the behavior of the physical system, selecting control or planning actions according to a formally defined objective. However, while such decisions are often optimal in a mathematical sense, their recommendations frequently diverge from human expectations. Although they may minimize costs, emissions or delays, their outputs often remain inequitable, opaque, or simply ``unnatural'' to the human thought ~\citep{stary2021digital,schink2026transforming}.

This divergence reflects an epistemic gap between the complete-information assumptions of algorithmic rationality and the heuristic, context-dependent reasoning characteristic of human bounded rationality~\citep{blandin2024learning,gowaikar2025efficiency}. As a result, recent research has increasingly sought to integrate these human-centered aspects into optimization frameworks (see, e.g.,~\citep{han2015formulation,chen2024optimal,villa2025fair,joo2019formalizing}). 

A common remedy has been to design handcrafted utility functions intended to approximate human values. However, such formulations are extremely cumbersome to construct and calibrate. As a concrete example,~\citep{pinto2025optimal} proposes an optimization-based framework to determine the placement of nature-based interventions (e.g., parks, trees, and green walls). In their formulation, the mixed-integer program’s objective coefficients must encode social, environmental, and aesthetic acceptability. Since these valuations are not analytically defined, the authors hand-tuned convolutional-like kernel weights, trading environmental indicators against perceived social impact. Clearly, this procedure is error-prone, non-repeatable, and highly domain-sensitive: two solutions that are equally ``optimal'' in the MILP sense may be perceived very differently by humans, and nothing in the formulation can express such preference gaps.

Other approaches make use of preference-based optimization~\citep{previtali2023glisp} and reinforcement learning from human feedback~\citep{christiano2017deep}, and offer a promising alternative to capture human expectation into decision engines. These approaches have been effectively used to align large language models to human preferences~\citep{rafailov2024scaling}, improve autonomous driving systems~\citep{nagahama2020autonomous}, and tune control policies in cyber-physical systems~\citep{zhu2023learning}. Collectively, these ideas signal a paradigm shift: objectives are no longer \emph{designed}, but \emph{elicited}. However, how such data-driven preference models can be reintegrated into optimization frameworks to explicitly capture ethical considerations remains largely unexplored~\citep{quaresmini2025role,villa2025epistemic}.

In this context, rather than viewing fairness (that is the latent context-dependent preference structure that humans implicitly apply when judging alternative decisions, i.e., the set of decision trade-offs that people consistently perceive as equitable or acceptable in a given situation) as an externally imposed constraint, we interpret it as an emergent property of bounded rational reasoning that the system must learn directly, without intermediaries who might potentially introduce bias.
This reframing has several implications. First, a fair DT must learn what ``good'' means in a specific context. Second, it emphasizes the importance of procedural fairness, where humans remain in the loop and iteratively guide the digital model through their feedback. Finally, the resulting decision problems must remain computationally efficient; user preferences must therefore be modeled through representations that are tractable and compatible with modern numerical optimization solvers~\citep{agrawal2021learning,abdufattokhov2021learning}. 

Building upon these insights, we propose a data-driven methodology that learns fairness as a latent objective inferred from pairwise user preferences. Starting from a classical optimization model, the DT explores feasible decisions and queries human evaluators (or simulated surrogates) to indicate which options they find more acceptable. By maintaining convexity, the learned cost can be seamlessly reinserted into the original solution workflow with little-to-no additional computational cost, yielding decisions that are both optimal and socially aligned.

\textbf{\emph{Contribution.}} This work advances fairness-aware Digital Twin design from theoretical framing to practical implementation. Specifically, we provide a complete end-to-end workflow that learns and embeds human-aligned objectives within classical optimization models. More precisely, we address what follows.

\begin{enumerate}
    \item \emph{Preference-driven learning workflow.} We develop a data-driven methodology that infers latent fairness objectives directly from human pairwise preferences over feasible solutions.

    \item \emph{Neural architecture for convex cost generation.} We introduce a novel Siamese neural network that predicts the parameters of a convex quadratic cost function conditioned on user preferences, ensuring tractability and solver compatibility.

    \item \emph{Real-world demonstrator.} We validate the proposed approach on a Digital Twin for COVID-19 hospital resource allocation, illustrating how the learned surrogate objectives improve social alignment without compromising efficiency.
\end{enumerate}

\textbf{\emph{Paper organization.}} The remainder of this paper is organized as follows. 
Section~\ref{sec:learning} formalizes the proposed preference-based framework for learning surrogate objectives from human feedback. 
Section~\ref{sec:nerualArchitecture} is dedicated to present our novel neural architecture. 
Section~\ref{sec:useCase} presents a proof-of-concept Digital Twin demonstrator for hospital resource allocation, illustrating the method’s practical implementation. 
Finally, Section~\ref{sec:conclusion} summarizes the findings and outlines directions for future research on fairness-aware Digital Twins.

\section{Learning a Surrogate Objective from Preferences}
\label{sec:learning}
Regardless of its domain, designing a Digital Twin always requires a \emph{decision engine} that selects the best action to perform on the actual system according to some user-defined criteria.  
This engine typically solves an optimization problem of the form:
\begin{subequations}
\label{eq:base_problem}
\begin{align}
&\min_{u} && f(u,x_0) \\
&\mathrm{s.t.} && g(u,x_0) \le 0 ,
\end{align}
\end{subequations}
where $u\in\mathbb{R}^{n_u}$ are the decision variables (actions) and
$x_0\in\mathbb{R}^{n_x}$ represents the current system context or feedback.
The function $f:\mathbb{R}^{n_u}\!\times\!\mathbb{R}^{n_x}\!\to\!\mathbb{R}$
encodes the objective to be minimized, while
$g:\mathbb{R}^{n_u}\!\times\!\mathbb{R}^{n_x}\!\to\!\mathbb{R}$ captures the
operational constraints.  The optimal decision $u^\star$ is then executed on
the physical system.

Although this formulation is mathematically well-posed, it assumes that the
objective $f$ is explicitly known.  In many practical scenarios, especially
those involving fairness, comfort, or social acceptability, such an explicit form is however unavailable or very difficult to be obtained, an operation that relies almost entirely on (potentially biased) human operators. 

\subsection{Preference-Based Cost Learning}

To address this issue, we propose a data-driven reformulation where the cost function itself is \emph{learned from human feedback}. Instead of explicitly prescribing what the system should optimize, we elicit user preferences between feasible alternatives and infer a latent objective function that, once placed into an appropriate optimization process, would reproduces those preferences.
To do so, starting from an instance of~\eqref{eq:base_problem}, we first sample a set of feasible decisions:
\begin{equation}
\mathcal{D}_0=\{(x_{0,i},u_i)\}_{i=1}^N,
\end{equation}
where each pair $(x_{0,i},u_i)$ denotes a feasible context–action combination.
User feedback is then collected through pairwise queries of the form:
\begin{quote}
For context $x_0$, which action do you prefer: $u_A$ or $u_B$?
\end{quote}
In many applications, the user can also be informed of the nominal objective
values $f(u_A,x_0)$ and $f(u_B,x_0)$, which often represent a measurable performance indicator such as cost, delay, or negative utility.
In the following, we assume that we fall such setting.

The resulting augmented dataset
\begin{equation}
\mathcal{D}=\{(x_{0,i},u_{A,i},u_{B,i},\ell_i)\}_{i=1}^N
\end{equation}
records the empirical preference label $\ell_i \in \{-1,+1\}$, where
$\ell_i=+1$ indicates that $u_{A,i}$ is preferred to $u_{B,i}$, that means $f(u_{A_i},x_0)<f(u_{B_i},x_0)$, and vice versa.

Given this data, we seek a cost parametric function $f(u,x_0) = f_\theta(u,x_0)$, where $\theta = \{\theta_i\}$ denotes the set of learnable parameters that characterize the function modeling human preference.  To this aim, in the remainder of the paper, we will exploit neural networks. In particular, we consider $f_\theta(u,x_0)$ to be a neural network where $\theta_i$ are the weights of the network to be identified. In the next section, we show how $f_\theta$ can be learned in a quadratic fashion.
Formally, this learning problem can be posed as the
\emph{Rank-SVM-like program}
\begin{subequations}
\label{eq:rank_svm}
\begin{align}
&\min_{\theta,\{\varepsilon_i\}}
&&\sum_{i=1}^N \varepsilon_i + \mathcal{R}(f_\theta)\\
&\text{s.t.} &&
\big(f_\theta(u_{A,i},x_{0,i}) - f_\theta(u_{B,i},x_{0,i})\big)\,\ell_i
\le -\delta + \varepsilon_i,\\
&&&\varepsilon_i \geq 0,\ \forall i,
\end{align}
\end{subequations}
where the regularizer $\mathcal{R}(f_\theta)$ promotes smoothness or
convexity, and the slack variables $\varepsilon_i\!\ge\!0$ absorb occasional
inconsistencies in human judgments.
The margin parameter $\delta\!>\!0$ enforces that the preferred action must be
better by a positive gap rather than by equality, preventing degenerate solutions that merely tie the two alternatives.

Once identified, the learned surrogate $f_\theta$ is used to parameterize~\eqref{eq:base_problem} (or replacing the original, biased $f$), yielding a
\emph{preference-aware optimization model}:
\begin{subequations}
\label{eq:preference_problem}
\begin{align}
&\min_{u} && f_\theta(u,x_0) \\
&\mathrm{s.t.} && g(u,x_0)\le 0.
\end{align}
\end{subequations}

\section{Neural Realization and Training}
\label{sec:nerualArchitecture}
While~\eqref{eq:rank_svm} defines the desired identification problem, solving it directly with standard deep learning toolboxes is difficult.
To lessen this issue, we relax the constrained formulation into its differentiable empirical-risk counterpart by replacing each inequality with a \emph{hinge ranking loss}
\begin{equation}
\mathcal{L}_{\mathrm{rank}}
= \sum_{i=1}^N \max\bigl(0,\,
f_\theta(u_{A,i},x_{0,i}) - f_\theta(u_{B,i},x_{0,i}) + \delta
\bigr),
\label{eq:hinge_loss}
\end{equation}
and minimize its empirical expectation over the dataset via stochastic gradient
descent.  This loss softly enforces the inequality
$f_\theta(u_A,x_0)<f_\theta(u_B,x_0)-\delta$
whenever $A\succ B$. To handle the other case $B\succ A$, we post-process $\mathcal{D}$ so that the preferred component is always the first one in the tuple.

To guarantee computational tractability of the downstream optimization, the model $f_\theta$ is not learned as a generic neural network but is shaped to produce a convex quadratic objective of the form
\begin{equation}
f_\theta(u,x_0)=\tfrac{1}{2}u^\top H(x_0)u+q(x_0)^\top u ,
\label{eq:quadratic_form}
\end{equation}
where the entries of $H(x_0)\succeq 0 $ and $q(x_0)$ are the output of a standard deep neural network. In particular, the NN provides as output the set of parameters $\theta = \{\theta_i\}_{i=1}^{n^2+n}$, and the first $n^2$ parameters fill the matrix $H$, while the last $n$ parameters fill $q$ (please refer to~\citep{abdufattokhov2021learning} for further details on the approach).  
This choice guarantees that the optimization problem
\eqref{eq:preference_problem}, as $x_0$ is fixed, can be solved with standard quadratic (mixed-integer) solvers without altering the constraint
structure or the solver stack and possibly preserving convexity.

To train $f_\theta(u,x_0)$, we employ the well-known
\emph{Siamese training scheme}: two identical copies of the same network $f_\theta(\cdot,x_0)$ are evaluated~\citep{chicco2021siamese}, respectively,
on the paired actions $(u_A,x_0)$ and $(u_B,x_0)$, and trained through the
hinge ranking loss~\eqref{eq:hinge_loss}.  
This architecture is not a different model class but a numerical device that
computes both sides of each preference pair through the same parameterized
function, ensuring that learning directly enforces the desired pairwise
ordering.

The learnt network does not predict the cost value directly.  Instead, for each context $x_0$, it will output the parameters of the quadratic cost:  
(i) a factor $S(x_0)$ whose product with its
transpose defines the Hessian $H(x_0)=S(x_0)S(x_0)^\top$, and  
(ii) the linear term $q(x_0)$.  
This structural design preserves convexity by construction and ensures that the
learned objective can be seamlessly reinserted into the solver in canonical quadratic form.  
The non-uniqueness of the factorization (any $S(x_0)Q$ with orthogonal $Q$
yields the same $H(x_0)$) introduces flat directions in parameter space, which
facilitates optimization by allowing multiple equally valid minima and prevents
the model from being trapped in sharp, isolated basins.

After training, querying the model for a new context $x_0$ yields a surrogate
objective
$$
f_\theta(u,x_0)
=\tfrac{1}{2}\,u^\top H(x_0)u + q(x_0)^\top u,
$$
whose level sets reflect empirically observed preferences while maintaining
solver compatibility and convexity guarantees.

\begin{remark}
Although input–convex neural architectures (ICNNs) could in principle learn
more general convex surrogates objectives, optimizing over these functions would make~\eqref{eq:preference_problem} fully non-linear. While this may be manageable in a continuous setting, it becomes highly problematic in mixed-integer formulations, where efficient MINLP solvers are rare and often unreliable. Our parametrization is therefore intentionally designed to avoid this issue.
\end{remark}

\subsection{Reducing Training Complexity via PCA Compression}
\label{sec:sec:pca_reduction}
Predicting a full Hessian $H(x_0)\!\in\!\mathbb{R}^{d\times d}$ directly in the original decision space is often computationally inefficient.  
Even for moderately sized problems, $d$ is in the order of hundreds or hundreds, implying $O(d^2)$ trainable parameters for the Hessian alone.

To mitigate this, we adopt a purely \emph{training-phase} dimensionality
reduction step.  
The decision vectors $u$ are projected onto a low-dimensional latent space
$$
z = W^\top (u - \mu),
$$
obtained via PCA on the collected feasible solutions.  
The Siamese model is then trained to produce a quadratic surrogate in the
compressed coordinates:
$$
f_\theta(z,x_0)=\tfrac{1}{2}z^\top H_z(x_0)\,z + q_z(x_0)^\top z .
$$
This reduces the number of trainable cost parameters from $O(d^2)$ to
$O(r^2)$, with $r\ll d$, yielding a substantial gain in sample efficiency,
training stability, and generalization.

Crucially, PCA is \emph{not} used to alter the final optimization domain.
After training, the learned quadratic is \emph{reconstructed} in the original
space via the analytic mapping
$$
H_u = W H_z W^\top, 
\qquad
q_u = W q_z - H_u\,\mu,
$$
so that the surrogate objective re-enters the original constrained problem
exactly as in~\eqref{eq:preference_problem}.  
All feasibility constraints, integer decisions, and operational couplings are therefore enforced in the full decision space, not in the PCA coordinates.

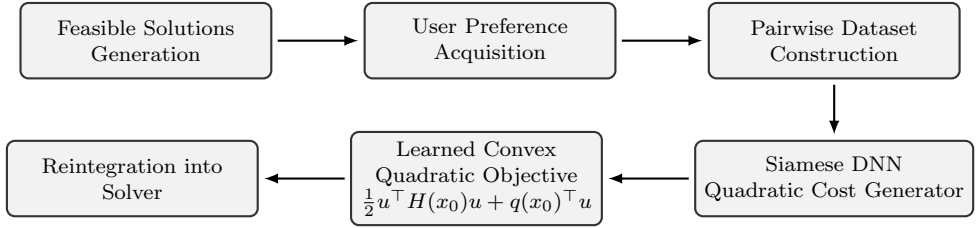
\begin{figure*}[t]
	\centering
	\begin{tikzpicture}[
		node distance=0.8cm and 1.2cm,
		every node/.style={font=\small, align=center},
		box/.style={rectangle, rounded corners=3pt, draw=black!80, thick, fill=gray!10, minimum width=3.3cm, minimum height=1.0cm},
		arrow/.style={-latex, thick, shorten >=2pt, shorten <=2pt}
		]
		
		\node[box] (bias) {Feasible Solutions \\ Generation};
		\node[box, right=of bias] (pref) {User Preference \\ Acquisition};
		\node[box, right=of pref] (pairs) {Pairwise Dataset \\ Construction};
		\node[box, below=of pairs] (siamese) {Siamese DNN \\ Quadratic Cost Generator};
		\node[box, left=of siamese] (qp) {Learned Convex \\ Quadratic Objective \\ $ \tfrac{1}{2}u^\top H(x_0)u + q(x_0)^\top u$};
		\node[box, left=of qp] (reuse) {Reintegration into \\ Solver};
		
		\draw[arrow] (bias) -- (pref);
		\draw[arrow] (pref) -- (pairs);
		\draw[arrow] (pairs) -- (siamese);
		\draw[arrow] (siamese) -- (qp);
		\draw[arrow] (qp) -- (reuse);

	\end{tikzpicture}
	\caption{Schematic overview of the demonstrator workflow. 
		Feasible suboptimal solutions are generated via biased optimization,
		scored by a simulated user function, and used to train a Siamese neural
		network that learns a convex quadratic objective consistent with user preferences.%
        }
	\label{fig:workflow}
\end{figure*}

The overall workflow of the demonstrator is summarized in Fig.~\ref{fig:workflow}.  

It comprises five main stages: (i) generation of diverse feasible decisions via biased optimization; (ii) scoring of these solutions through a simulated user preference model; (iii) construction of a pairwise preference dataset; (iv) learning of a surrogate convex quadratic objective; and (v) reintegration of the learned objective into the original solver for fairness-aware optimization.

\section{Proof-of-Concept Demonstrator: COVID-19 Facility Allocation}

\label{sec:useCase}
As a representative constrained decision problem, we consider the
COVID-19 healthcare facility-allocation benchmark distributed by Gurobi\footnote{\url{https://github.com/Gurobi/modeling-examples/tree/master/covid19_facility_location}},
in which patients from multiple counties must be assigned to existing or
temporary medical facilities.  The problem naturally fits the Digital Twin
paradigm: the virtual decision engine must repeatedly propose allocations
under changing conditions, balancing efficiency with perceived acceptability.

\subsection{Mathematical Model}

Let $\mathcal{C}=\{1,\dots,9\}$ denote counties (demand nodes) and
$\mathcal{F}=\{1,\dots,23\}$ the set of available facilities, partitioned
into existing $\mathcal{F}_E$ and temporary $\mathcal{F}_T$ units.
For each $(c,f)\in\mathcal{C}\!\times\!\mathcal{F}$, let
$x_{cf}\ge 0$ denote the number of patients from county $c$ assigned
to facility $f$.  Activation of temporary facilities is modeled via binary
variables $y_f\in\{0,1\}$ for $f\in\mathcal{F}_T$.
The forecast demand of county $c$ is $D_c$, the capacity of facility $f$
is $\mathrm{cap}_f$, and the cost of assigning one patient from $c$ to $f$
is $d_{cf}$ (proportional to travel distance).
Opening a temporary facility incurs a fixed cost $K_f$.

The base allocation problem reads:
\begin{subequations}
\label{eq:facility_allocation}
\begin{align}
\min_{x,y}\quad &
\sum_{c\in\mathcal{C}}\sum_{f\in\mathcal{F}} d_{cf}\,x_{cf}
\;+\;
\sum_{f\in\mathcal{F}_T} K_f\,y_f
\\[0.4em]
\text{s.t.}\quad &
\sum_{f\in\mathcal{F}} x_{cf} = D_c,
&& \forall c\in\mathcal{C}
\\
&
\sum_{c\in\mathcal{C}} x_{cf} \le \mathrm{cap}_f,
&& \forall f\in\mathcal{F}_E
\\
&
\sum_{c\in\mathcal{C}} x_{cf} \le \mathrm{cap}_f\,y_f,
&& \forall f\in\mathcal{F}_T
\\
&
x_{cf}\ge 0, && \forall (c,f), \\
& y_f\in\{0,1\},\ \forall f\in\mathcal{F}_T .
\end{align}
\end{subequations}

This is a mixed-integer linear program combining transportation costs and
capacity constraints under activation decisions.  In the following, we use this model to showcase our approach.

\subsection{Benchmark Model and Feasible Solution Generation}

To generate a variety of feasible yet distinct assignments, we perturb the nominal objective with a quadratic bias term:
$$
\min_{x,y} J_{\text{orig}}(x,y)
+ \lambda \big(\|x - x^{\text{bias}}\|_F^2 + \|y - y^{\text{bias}}\|_2^2 \big),
$$
where $(x^{\text{bias}},y^{\text{bias}})$ are randomly sampled reference configurations around the baseline optimum, and $\lambda$ controls the bias intensity.  
Each biased optimization run yields a feasible, suboptimal solution reflecting a different operational trade-off--such as prioritizing specific hospitals or minimizing load imbalance.  
This process results in a diverse collection of feasible policies $\{(x_i,y_i)\}$ that all satisfy the system constraints but differ in cost and structure.
\subsection{Simulated User Preference Scoring}

To emulate human evaluation in a controlled setting, each feasible solution is
assigned a synthetic \emph{user preference} score in a 5-dimensional {feature space} derived from the high-dimensional solution capturing qualitative
acceptability beyond pure efficiency.

Each assignment matrix $x_i$ is mapped to a five-dimensional feature vector
$\bar u_i$ summarizing interpretable system characteristics:
(1) total load on the first existing facility,
(2) total load on the last facility,
(3) total load on temporary facilities,
(4) maximum county-to-facility assignment, and
(5) overall demand served (all normalized).
These five features act as a compressed representation of the decision's
high-dimensional structure.

The feature vector is then evaluated through a multidimensional Rosenbrock function
\begin{equation}
R(\bar  u) = \sum_{k=1}^{4} \bigl[100(\bar  u_{k+1}-u_k^2)^2 + (1-\bar  u_k)^2\bigr],
\label{eq:user_prefOracle}
\end{equation}
used here to mimic complex, non-monotonic human preference
patterns. Lower values correspond to more desirable decisions.
To model context dependence, a scalar $x_0 \in [0,1]$ is introduced and
the feature vector is shifted as $\bar u' = \bar u - 0.1x_0$, representing how the same allocation may be perceived differently under changing conditions (e.g., pandemic intensity or resource availability).

In the demonstrator, preferences are not derived from $s_i$ alone: to reflect that users also value nominal performance, the ranking signal is constructed using the
\emph{composite score}
\[
\phi_i := s_i + J_\text{orig}(y_i,x_i),
\]
so that a solution with lower $\phi_i$ is deemed preferable.  When constructing
pairwise comparisons, we always order the pairs so that $u_A$ denotes the
preferred element, i.e., the one with strictly lower composite score
$\phi_A < \phi_B$.  As a consequence, all preference labels entering the
learning stage have the canonical form “$u_A$ is preferred to $u_B$” encoded implicitly by construction.

\subsection{Reintegration and Validation}

Once trained, the surrogate cost is reconstructed in the original decision
space and reintroduced into the base optimization
model:
\[
\min_{u}\ \tfrac{1}{2}u^\top H(x_0)\,u + q(x_0)^\top u
\qquad \text{s.t.}\qquad g(u,x_0)\le 0.
\]
Please note that the constraint function $g$ remains the original one, so feasibility is always guaranteed. 

Although the present prototype uses synthetic preference feedback, the loop is
agnostic to the supervision source: real human feedback (binary choices,
coarse ratings, or contextual scores) can be injected without modifying the
learning architecture or the downstream solver.

\subsection{Implementation Details}

The base optimization model is implemented in \texttt{Python} using \texttt{CVXPY} and the \texttt{SCIP} solver.  The overall problem comprises 216 decision variables, either continuous or binary.
The siamese network is implemented in \texttt{Keras} using a Siamese
architecture.  The network that generates the quadratic cost parameter is a feedforward MLP with three hidden layers of 20 neurons, equipped with \texttt{ReLU} activations, and $\ell_2$ regularization on weights and biases.
Before training, all decision vectors $u$ are projected onto a
30-dimensional latent space via PCA (30 is empirically set to balance training speed and accuracy).

Training is performed using the Adam optimizer, with a 10\% of the training set retained for a validation split, learning-rate scheduling, and early stopping. 
After convergence, the network is queried over all distinct context values to
export the corresponding $(H(x_0),q(x_0))$ pairs for reintegration into the
original optimization model.

To generate a finite set of feasible solutions, the nominal optimum is first computed, and additional feasible solutions are
obtained by perturbing the objective with randomly generated bias terms of the form
\[
x^{\text{bias}} = x^{\star} + \Delta_x,\qquad
\Delta_x \sim \mathcal{U}[-w,w],
\]
with an analogous perturbation for $y$, where $w>0$ controls the magnitude of
exploration around the optimum.

For training, the context parameter $x_0$ is discretized over a uniform grid of $26$ in $[0,1]$, and for each $x_0$ the above procedure yields a finite batch of feasible allocations.  
Given the feasible set under a fixed context, all possible ordered
preference pairs $(u_A,u_B)$ with distinct composite scores $\phi$ could in principle be considered.  In practice, to control dataset size and simulate information scarcity, we randomly subsample a finite number of such pairs.  
To emulate imperfect user judgments, a fraction of the generated labels is
optionally flipped; both the subsampling rate and the flip probability are
treated as hyperparameters and swept in the evaluation study.

The resulting dataset consists of context–dependent, noise–corrupted preference
pairs of the canonical form “$u_A$ is preferred over $u_B$ at context $x_0$”,
which are then used to fit the surrogate objective described above.
After training, for reintegration,  $x_0$ is instead discretized over a uniform grid of $52$ steps in $[0,1]$.

\subsection{Experimental Evaluation}

For each context value $x_0$, we solved the reintegration problem
\eqref{eq:preference_problem} with the learned parameters $(H(x_0),q(x_0))$
replacing the nominal objective, and we recorded a \emph{win} whenever the resulting allocation achieved a strictly lower composite cost, i.e.
$J_{\text{orig}}(x,y) + R(x,y,x_0)$
than the nominal solution under the same context.   The experiment was repeated over seven random seeds; the results were averaged.

Table~\ref{tab:surrogate_wins} reports the average number of wins (out of $52$
contexts) for different dataset sizes and controlled label corruption levels.
The surrogate consistently outperforms the nominal objective whenever preference
labels are mostly correct and sufficient data is provided.  Performance
degrades only when a large fraction of labels is flipped,
confirming that the observed gain is supervision-driven rather than an artifact
of regularization or overfitting.

\begin{table}[h!]
\centering
\caption{Average number of contexts (out of $52$) in which the surrogate objective yields a lower composite cost than the nominal solution, averaged
over seven seeds.}
\vspace{0.2em}
\begin{tabular}{c|cccc}
\toprule
\multirow{2}{*}{\textbf{Dataset size (pairs)}} &
\multicolumn{4}{c}{\textbf{Fraction of flipped labels}} \\
& 0\% & 10\% & 20\% & 30\% \\
\midrule
448  & 48.429 & 50.571 & 44.857 & 18.714 \\
896  & 51.143 & 51.429 & 47.429 & 39.571 \\
1344 & 52     & 52     & 41.429 & 47.143 \\
\bottomrule
\end{tabular}
\label{tab:surrogate_wins}
\end{table}

Beyond aggregate win statistics, it is interesting to visualize how the surrogate redistributes cost and preference score across contexts.
For a fixed configuration of training hyperparameters, we run multiple seeds
and, for each context $x_0$, compute the across–seed mean and standard
deviation of
\[
\Delta R := R_{\text{nom}} - R_{\text{sur}},\qquad
\Delta {J} := {J}_{\text{sur}} - {J}_{\text{nom}} .
\]
Thus $\Delta R>0$ indicates a gain in perceived acceptability, while
$\Delta \mathrm{cost}>0$ represents a penalty in nominal efficiency.

\begin{figure}[t]
\centering

\subfloat[448 samples and 10\% wrong labels\label{fig:A}]{
    \includegraphics[clip,trim={0 0 0 1.5cm},width=\linewidth]{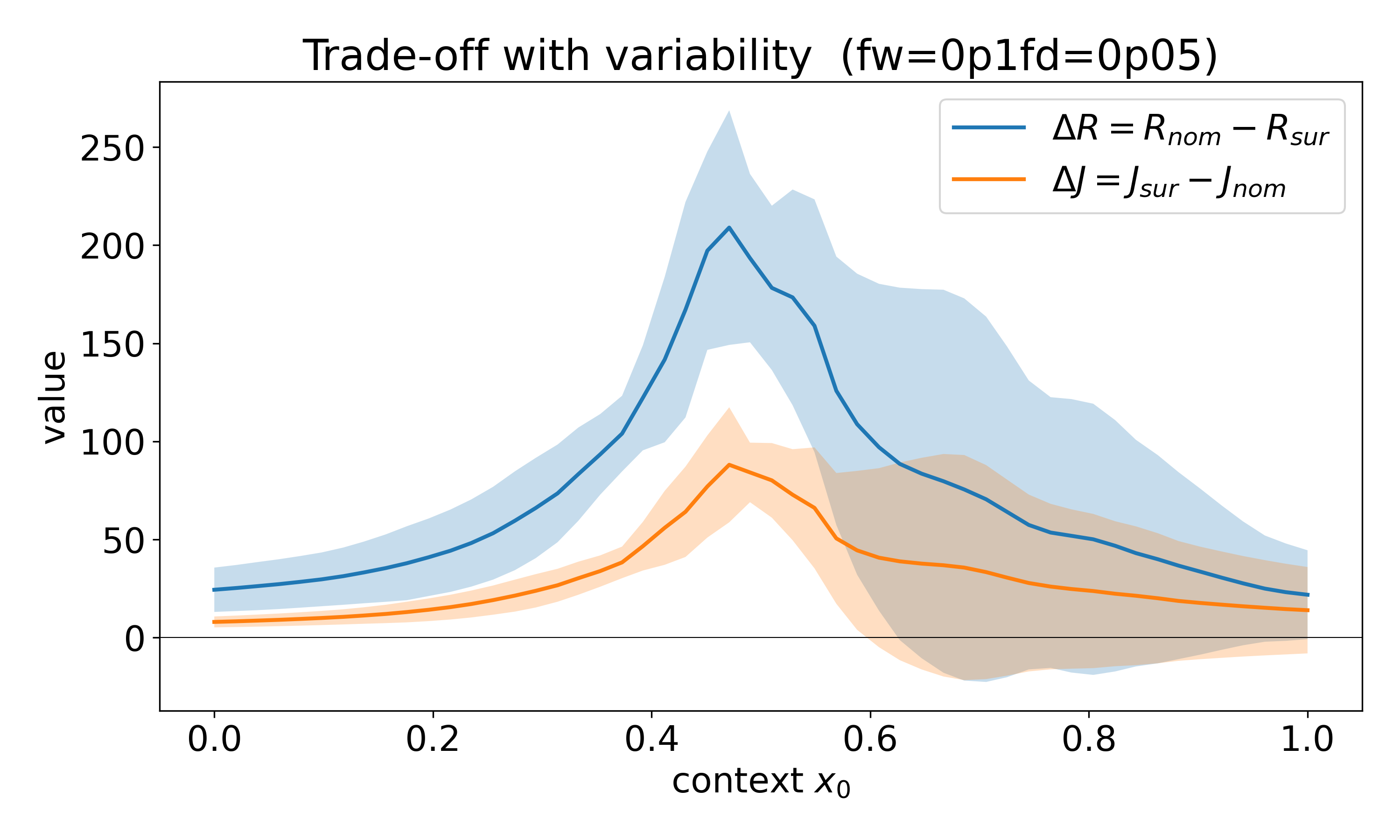}
}\vspace{0.4cm}

\subfloat[1344 samples and no wrong labels\label{fig:B}]{
    \includegraphics[clip,trim={0 0 0 1.45cm},width=\linewidth]{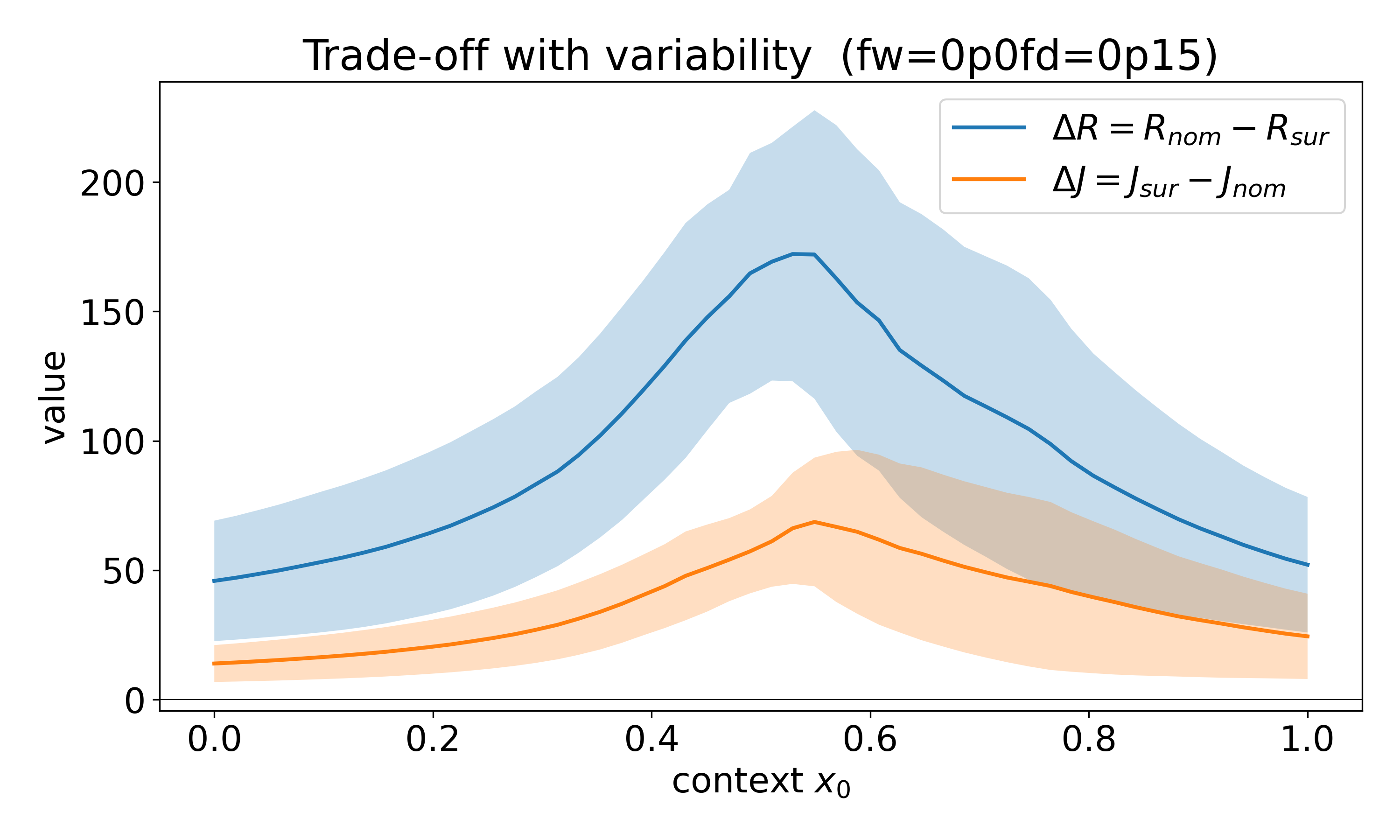}
}

\caption{Qualitative inspection of the surrogate–induced trade--off for two
training configurations $(f_w,f_d)$. For each context $x_0$, we plot the
mean $\pm$ standard deviation across seeds of
$\Delta R := R_{\mathrm{nom}} - R_{\mathrm{sur}}$ and
$\Delta J := J_{\mathrm{sur}} - J_{\mathrm{nom}}$.
Positive $\Delta R$ indicates a gain in perceived acceptability, while
positive $\Delta J$ indicates nominal--efficiency loss.}
\label{fig:tradeoff_multi}
\end{figure}
Panel~\ref{fig:A} corresponds to the smallest dataset with
10\% flipped labels.  The surrogate still achieves sizeable gains in perceived
acceptability (\(\Delta R>0\)) over a broad range of contexts, peaking around
mid–range \(x_0\), but the variability bands are wide and occasional sign
reversals appear near \(x_0\!\approx\!0.6\).  The nominal cost penalty
(\(\Delta J>0\)) rises with context and exhibits a pronounced hump, suggesting
that, under weak and noisy supervision, the learned objective trades efficiency
for acceptability in a less calibrated, context–dependent manner.
Panel~\ref{fig:B} uses the largest dataset with no label
corruption.  The pattern becomes markedly cleaner: \(\Delta R\) remains
consistently positive with tighter bands, and the \(\Delta J\) curve is both
smoother and lower in magnitude relative to (a).  This indicates that abundant,
reliable feedback enables the surrogate to extract a more stable preference
signal and to realize comparable acceptability gains at a reduced efficiency
cost.
\section{Conclusion}
\label{sec:conclusion}

This paper proposed a data-driven framework to embed fairness in optimization-based Digital Twins through preference learning. We introduced a workflow that infers human-aligned surrogate objectives from pairwise preferences and realized it via a Siamese neural network that generates convex quadratic costs, ensuring tractability and solver compatibility. A proof-of-concept study on COVID-19 hospital resource allocation demonstrated that the learned objectives can improve perceived fairness without compromising nominal efficiency. 
Future work will extend this framework to scenarios involving multiple users with potentially diverging preferences, aiming to reconcile conflicting feedback signals and model collective fairness equilibria. Additional research will explore real human-in-the-loop and online learning scenarios.

\section*{Declaration of Generative AI and AI-assisted technologies in the
writing process}

During the preparation of this work, the authors used ChatGPT and Gemini in
order to improve readability and language. After using these services,
the authors reviewed and edited the content as needed and take full
responsibility for the content of the publication.

\bibliography{biblio}

\begin{thebibliography}{22}
\providecommand{\natexlab}[1]{#1}
\providecommand{\url}[1]{\texttt{#1}}
\providecommand{\urlprefix}{URL }
\expandafter\ifx\csname urlstyle\endcsname\relax
  \providecommand{\doi}[1]{doi:\discretionary{}{}{}#1}\else
  \providecommand{\doi}{doi:\discretionary{}{}{}\begingroup
  \urlstyle{rm}\Url}\fi

\bibitem[{Abdufattokhov et~al.(2021)Abdufattokhov, Zanon, and
  Bemporad}]{abdufattokhov2021learning}
Abdufattokhov, S., Zanon, M., and Bemporad, A. (2021).
\newblock Learning convex terminal costs for complexity reduction in mpc.
\newblock In \emph{2021 60th IEEE Conference on Decision and Control (CDC)},
  2163--2168. IEEE.

\bibitem[{Agrawal et~al.(2021)Agrawal, Barratt, and Boyd}]{agrawal2021learning}
Agrawal, A., Barratt, S., and Boyd, S. (2021).
\newblock Learning convex optimization models.
\newblock \emph{IEEE/CAA Journal of Automatica Sinica}, 8(8), 1355--1364.

\bibitem[{Blandin and Kash(2024)}]{blandin2024learning}
Blandin, J. and Kash, I.A. (2024).
\newblock Learning fairness from demonstrations via inverse reinforcement
  learning.
\newblock In \emph{Proceedings of the 2024 ACM Conference on Fairness,
  Accountability, and Transparency}, 51--61.

\bibitem[{Chen et~al.(2024)Chen, Hu, Xie, Zheng, Hu, and
  Yang}]{chen2024optimal}
Chen, Y., Hu, S., Xie, S., Zheng, Y., Hu, Q., and Yang, Q. (2024).
\newblock Optimal dynamic pricing of fast charging stations considering bounded
  rationality of users and market regulation.
\newblock \emph{IEEE Transactions on Smart Grid}, 15(4), 3950--3965.

\bibitem[{Chicco(2021)}]{chicco2021siamese}
Chicco, D. (2021).
\newblock Siamese neural networks: An overview.
\newblock \emph{Artificial neural networks}, 73--94.

\bibitem[{Christiano et~al.(2017)Christiano, Leike, Brown, Martic, Legg, and
  Amodei}]{christiano2017deep}
Christiano, P.F., Leike, J., Brown, T., Martic, M., Legg, S., and Amodei, D.
  (2017).
\newblock Deep reinforcement learning from human preferences.
\newblock \emph{Advances in neural information processing systems}, 30.

\bibitem[{Djebali et~al.(2024)Djebali, Guerard, and Taleb}]{djebali2024survey}
Djebali, S., Guerard, G., and Taleb, I. (2024).
\newblock Survey and insights on digital twins design and smart grid’s
  applications.
\newblock \emph{Future Generation Computer Systems}, 153, 234--248.

\bibitem[{Gowaikar et~al.(2025)Gowaikar, Berard, Mushkani, and
  Koseki}]{gowaikar2025efficiency}
Gowaikar, S., Berard, H., Mushkani, R., and Koseki, S. (2025).
\newblock From efficiency to equity: Measuring fairness in preference learning.
\newblock In \emph{Proceedings of the AAAI/ACM Conference on AI, Ethics, and
  Society}, volume~8, 1134--1143.

\bibitem[{Han et~al.(2015)Han, Szeto, and Friesz}]{han2015formulation}
Han, K., Szeto, W.Y., and Friesz, T.L. (2015).
\newblock Formulation, existence, and computation of boundedly rational dynamic
  user equilibrium with fixed or endogenous user tolerance.
\newblock \emph{Transportation Research Part B: Methodological}, 79, 16--49.

\bibitem[{Joo and Shin(2019)}]{joo2019formalizing}
Joo, T. and Shin, D. (2019).
\newblock Formalizing human--machine interactions for adaptive automation in
  smart manufacturing.
\newblock \emph{IEEE Transactions on Human-Machine Systems}, 49(6), 529--539.

\bibitem[{Katsoulakis et~al.(2024)}]{katsoulakis2024digital}
Katsoulakis, E. et~al. (2024).
\newblock Digital twins for health: a scoping review.
\newblock \emph{NPJ digital medicine}, 7(1), 77.

\bibitem[{Nagahama et~al.(2020)Nagahama, Saito, Wada, and
  Sonoda}]{nagahama2020autonomous}
Nagahama, A., Saito, T., Wada, T., and Sonoda, K. (2020).
\newblock Autonomous driving learning preference of collision avoidance
  maneuvers.
\newblock \emph{IEEE Transactions on Intelligent Transportation Systems},
  22(9), 5624--5634.

\bibitem[{Pinto et~al.(2025)Pinto, Russo, and Sudoso}]{pinto2025optimal}
Pinto, D.M., Russo, D.D., and Sudoso, A.M. (2025).
\newblock Optimal placement of nature-based solutions for urban challenges.
\newblock \emph{arXiv preprint arXiv:2502.11065}.

\bibitem[{Previtali et~al.(2023)Previtali, Mazzoleni, Ferramosca, and
  Previdi}]{previtali2023glisp}
Previtali, D., Mazzoleni, M., Ferramosca, A., and Previdi, F. (2023).
\newblock \texttt{GLISp-r}: a preference-based optimization algorithm with
  convergence guarantees.
\newblock \emph{Computational Optimization and Applications}, 86(1), 383--420.

\bibitem[{Quaresmini et~al.(2025)Quaresmini, Villa, Maghool, Breschi,
  Schaffonati, and Tanelli}]{quaresmini2025role}
Quaresmini, C., Villa, E., Maghool, S., Breschi, V., Schaffonati, V., and
  Tanelli, M. (2025).
\newblock The role of epistemic fairness in dynamics models to support
  sustainable mobility diffusion.
\newblock In \emph{Engineering and Value Change}, 179--198. Springer.

\bibitem[{Rafailov et~al.(2024)Rafailov, Chittepu, Park, Sikchi, Hejna, Knox,
  Finn, and Niekum}]{rafailov2024scaling}
Rafailov, R., Chittepu, Y., Park, R., Sikchi, H.S., Hejna, J., Knox, B., Finn,
  C., and Niekum, S. (2024).
\newblock Scaling laws for reward model overoptimization in direct alignment
  algorithms.
\newblock \emph{Advances in Neural Information Processing Systems}, 37,
  126207--126242.

\bibitem[{Schink et~al.(2026)Schink, Gutmann, and
  Brenk}]{schink2026transforming}
Schink, A., Gutmann, T., and Brenk, S. (2026).
\newblock Transforming b2b platforms through interconnected digital twins:
  Enhancing situation awareness for decision-making.
\newblock \emph{Technovation}, 149, 103368.

\bibitem[{Somma et~al.(2025)Somma, Amalfitano, Bucaioni, and
  De~Benedictis}]{somma2025model}
Somma, A., Amalfitano, D., Bucaioni, A., and De~Benedictis, A. (2025).
\newblock A model-driven approach for engineering mobility digital twins: The
  bologna case study.
\newblock \emph{Information and Software Technology}, 107863.

\bibitem[{Stary(2021)}]{stary2021digital}
Stary, C. (2021).
\newblock Digital twin generation: Re-conceptualizing agent systems for
  behavior-centered cyber-physical system development.
\newblock \emph{Sensors}, 21(4), 1096.

\bibitem[{Villa et~al.(2025{\natexlab{a}})Villa, Breschi, and
  Tanelli}]{villa2025fair}
Villa, E., Breschi, V., and Tanelli, M. (2025{\natexlab{a}}).
\newblock {Fair-MPC}: A framework for just decision-making.
\newblock \emph{IEEE Transactions on Automatic Control}.

\bibitem[{Villa et~al.(2025{\natexlab{b}})Villa, Quaresmini, Breschi,
  Schiaffonati, and Tanelli}]{villa2025epistemic}
Villa, E., Quaresmini, C., Breschi, V., Schiaffonati, V., and Tanelli, M.
  (2025{\natexlab{b}}).
\newblock The epistemic dimension of algorithmic fairness: assessing its impact
  in innovation diffusion and fair policy making.
\newblock \emph{arXiv preprint arXiv:2504.02856}.

\bibitem[{Zhu et~al.(2023)Zhu, Bemporad, Kneissl, and Esen}]{zhu2023learning}
Zhu, M., Bemporad, A., Kneissl, M., and Esen, H. (2023).
\newblock Learning critical scenarios in feedback control systems for automated
  driving.
\newblock In \emph{2023 IEEE 26th International Conference on Intelligent
  Transportation Systems (ITSC)}, 321--328. IEEE.

\end{thebibliography}
\end{document}